\pdfoutput=1
\documentclass[11pt]{article}
\usepackage{acl}

\usepackage{times}
\usepackage{latexsym}
\usepackage[T1]{fontenc}
\usepackage[utf8]{inputenc}
\usepackage{microtype}
\usepackage{inconsolata}
\usepackage{graphicx}
\usepackage{array}
\usepackage{amsmath}
\usepackage{amssymb}

\begin{document}

\title{{LLM}-{S}upported {N}atural {L}anguage to {B}ash {T}ranslation}
\author{
  Finnian Westenfelder$^{1,2}$, Erik Hemberg$^1$, Miguel Tulla$^1$,\\
  \bf Stephen Moskal$^1$, \bf Una-May O'Reilly$^1$, \bf Silviu Chiricescu$^3$\\
  $^1$ALFA Group MIT-CSAIL, $^2$Draper Scholar, $^3$Charles Stark Draper Laboratory\\
  \texttt{\{finnw,ehemberg,mtulla,smoskal,unamay\}@mit.edu silviu@draper.com}
}

\maketitle

\begin{abstract}
  The Bourne-Again Shell (Bash) command-line interface for Linux systems has complex syntax and requires extensive specialized knowledge. Using the natural language to Bash command (NL2SH) translation capabilities of large language models (LLMs) for command composition circumvents these issues. However, the NL2SH performance of LLMs is difficult to assess due to inaccurate test data and unreliable heuristics for determining the functional equivalence of Bash commands. We present a manually verified test dataset of 600 instruction-command pairs and a training dataset of 40,939 pairs, increasing the size of previous datasets by 441\% and 135\%, respectively. Further, we present a novel functional equivalence heuristic that combines command execution with LLM evaluation of command outputs. Our heuristic can determine the functional equivalence of two Bash commands with 95\% confidence, a 16\% increase over previous heuristics. Evaluation of popular LLMs using our test dataset and heuristic demonstrates that parsing, in-context learning, in-weight learning and constrained decoding can improve NL2SH accuracy up to 32\%. Our findings emphasize the importance of dataset quality, execution-based evaluation and translation method for advancing NL2SH translation. Our code is available at \href{https://github.com/westenfelder/NL2SH}{https://github.com/westenfelder/NL2SH}.
\end{abstract}

\section{Introduction}
\label{sec:introduction}
The default command-line interface (CLI) for interacting with Linux systems is the Bourne-Again Shell (Bash) \cite{linux_cli_book}. Bash commands allow computer users to control processes, interact with the file system and manage the network. However, using Bash requires knowledge of numerous utilities, each with unique parameters and complex syntax \cite{Bash}. Moreover, the reference documentation for these utilities, called manual pages, can be cumbersome and confusing \cite{manpages}. This makes the CLI a barrier for inexperienced users and increases the chance of errors for experienced users \cite{NL2CMD-Comp}.

Language models that convert natural language to command-line instructions, a task referred to as \textit{NL2SH, NL2CMD or NL2Bash Translation}, offer a promising solution to this problem \cite{NL2CMD-Comp}. We use the term \textit{NL2SH model} to refer to models trained specifically for the task of NL2SH, as well as the NL2SH capabilities of general-purpose large language models (LLMs). Figure \ref{fig:example} shows an example of natural language to Bash command translation. NL2SH models are well suited for CLIs because they are designed for text-based interactions.  NL2SH models can simplify human-computer interactions by allowing users to interact with Linux systems through natural language on the command line. This advancement enhances usability by reducing the need for syntax memorization \cite{Sammet}.

\begin{figure}[ht!]
  \centering
  \small
  \begin{tabular}{l}
    \hline
    \textbf{Input:} Natural Language                     \\
    \textit{List files in the /workspace directory that} \\
    \textit{were accessed over an hour ago.}             \\
    \hline
    \textbf{Output:} Bash Command                        \\
    \texttt{find /workspace -type f -amin +60}           \\
    \hline
  \end{tabular}
  \caption{Natural language to Bash command translation example from our NL2SH-ALFA dataset.}
  \label{fig:example}
\end{figure}

The use of NL2SH models necessitates benchmarks to measure task accuracy \cite{warp, shell-gpt, copilot, msai, cwai}. A NL2SH benchmark requires test data consisting of natural language prompts and ground truth commands (referred to as instruction-command pairs). Given a natural language prompt, a NL2SH model generates a Bash command. A benchmark must then use a heuristic to determine if the model command is functionally equivalent to the ground truth command. Determining functional equivalence of commands is difficult because there are multiple possible correct commands for a given task, due to a wide range of interchangeable utilities. Further, command execution may not result in identical outputs, neutralizing evaluation with string comparison. Current benchmarks do not accurately measure NL2SH model performance due to errors in assessment data and inaccurate heuristics for determining the functional equivalence of commands \cite{InterCode,tsed,codesift}. This makes it difficult to assess model capabilities and measure methods for improving model performance.

To address these challenges, we investigate the following research questions: (1) How can we validate NL2SH datasets to ensure models are evaluated using accurate assessments? (2) How can we design a functional equivalence heuristic that accurately measures the quality of model translations? (3) How can we improve the accuracy of NL2SH models as measured by a reliable benchmark?

Our contributions are summarized as follows: (1) We create a manually verified test dataset of 600 instruction-command pairs and a training dataset of 40,939 pairs, increasing the size of previous test and training datasets by 441\% and 135\%, respectively. (2) We present a novel functional equivalence heuristic that combines command execution with LLM evaluation of command outputs, capable of determining the functional equivalence of two Bash commands with 95\% confidence, a 16\% increase over previous heuristics. (3) We evaluate popular LLMs using our test data and heuristic and demonstrate that parsing, in-context learning, in-weight learning and constrained decoding can improve NL2SH accuracy by up to 32\%.

\section{Background}
\label{sec:background}
NL2SH translation falls under the broader domain of machine translation, where models automatically translate text or speech from one language to another. LLMs are well suited for this task, enabling translation that was impossible with previous methods \cite{machinetranslation}. Evaluating NL2SH models requires determining the functional correctness of generated commands. Functional correctness is defined as whether the code produces the correct output for each input, as specified, or as compared to ground truth \cite{codecorrectness}. Functional correctness does not consider the diversity of code generated, or other factors such as run time and memory consumption \cite{funccorrectenough}.

Ensuring the functional correctness of code is difficult because validation methods are error-prone and take an impractical amount of time for large volumes of code \cite{codecorrectness}. There are two main validation techniques: static and dynamic analysis. Static analysis checks code without execution, using parsers, lexical analysis or control flow checking \cite{codepatch}. Dynamic analysis evaluates code outputs and runtime behavior using an execution environment \cite{InterCode}. Some frameworks combine static and dynamic analysis \cite{codesift}.

Determining the functional correctness of Bash commands translated from natural language is a sub-problem of validating code correctness. We assess the functional correctness of Bash commands by comparing the generated (model) command with a ground truth Bash command. We define the term \textit{"functional equivalence heuristic" (FEH)} to describe a heuristic that performs this comparison and determines the functional correctness of a Bash command. Due to varying definitions in this field, we provide notation in Table \ref{tab:notation} defining the terms used in this paper.

\begin{table}[ht!]
  \centering
  \scriptsize
  \caption{Definition of terms.}
  \renewcommand{\arraystretch}{1.3}
  \begin{tabular}{ll}
    \hline
    \textbf{Term}           & \textbf{Definition}                                                           \\
    \hline
    Natural Language Task   & \( t \in T \) English                                                         \\
    \hline
    Ground Truth Command    & \( b \in B \) Bash-5.2                                                        \\
    \hline
    Model Command           & \( b' \in B \)                                                                \\
    \hline
    Functionally Equivalent & \( \hat{b} \in B \)                                                           \\
    Command                 &                                                                               \\
    \hline
    Docker Environment      & \( e \in E \) Linux                                                           \\
    \hline
    Command Output          & \( o \in O \) stdout and system state                                         \\
    and Side Effects        &                                                                               \\
    \hline
    Model Weights           & \( \theta \in \mathbb{R} \)                                                   \\
    \hline
    Translation             & \( f: T \times \mathbb{R} \rightarrow B \quad f(t, \theta) = b' \)            \\
    \hline
    Execution               & \( g: B \times E \rightarrow O \)                                             \\
                            & \( g(b, e) = o \quad g(b', e) = o' \)                                         \\
    \hline
    Ideal FEH               & \( m: T \times O \times O \rightarrow \{0,1\} \)                              \\
                            & \( m(t, o, o') = \begin{cases} 1: o \approx o' \\ 0: o \neq o' \end{cases} \) \\
    \hline
    Training Dataset        & \( D_T: \{(t_i, b_i)\ |\ i = 1, 2, \dots, x\} \)                              \\
    \hline
    Test Dataset            & \( D_H: \{(t_i, b_i, \hat{b_i})\ |\ i = 1, 2, \dots, y\} \)                   \\
                            & \( m(t, g(b, e), g(\hat{b}, e)) = 1 \quad \text{ideal } m \)                  \\
                            & \( D_T \cap D_H = \emptyset \)                                                \\
    \hline
    Benchmark               & \( (D_H, m) \)                                                                \\
    \hline
  \end{tabular}
  \label{tab:notation}
\end{table}

\vspace{-12pt}

\section{Related Work}
\label{sec:related-work}
NL2SH translation is a well-studied natural language processing (NLP) task. Table \ref{tab:related} summarizes the contributions of previous work by listing the names of datasets, functional equivalence heuristics (FEHs), and models used in this field. The table is sparsely populated because the majority of previous work focused on improving a NL2SH dataset, FEH, or model in isolation.

\begin{table}[ht!]
  \scriptsize
  \setlength{\tabcolsep}{2pt}
  \centering
  \caption{Summary of NL2SH datasets, FEHs, and models created in previous work.}
  \begin{tabular}{llll}
    \hline
    \textbf{Citation}       & \textbf{Datasets} & \textbf{FEHs} & \textbf{Models}  \\ \hline
    \citet{NL2Bash}         & NL2Bash           & -             & -                \\
    \citet{AInix}           & -                 & -             & AInix            \\
    \citet{NL2CMD-Comp}     & -                 & NL2CMD        & Tellina          \\
    \citet{Magnum}          & -                 & -             & Magnum           \\
    \citet{dataset2}        & NL2CMD            & -             & -                \\
    \citet{AST}             & -                 & -             & AST              \\
    \citet{finetune}        & -                 & -             & T5, GPT2         \\
    \citet{ShellGPT}        & -                 & -             & ShellGPT         \\
    \citet{InterCode}       & InterCode-Bash    & InterCode     & -                \\
    \citet{dataset4}        & text\_to\_bash    & -             & -                \\
    \citet{multiple}        & MultiPL-E         & Unit Tests    & -                \\
    \citet{tsed}            & -                 & TSED          & -                \\
    \citet{codesift}        & CodeSift          & CodeSift      & -                \\
    \citet{exec_based_eval} & IBM\_Instana      & Podman        & -                \\
    \citet{dataset3}        & LinuxCmds         & -             & -                \\
    \citet{warp}            & -                 & -             & Warp AI          \\
    \citet{shell-gpt}       & -                 & -             & shell-gpt        \\
    \citet{copilot}         & -                 & -             & Copilot CLI      \\
    \citet{cwai}            & -                 & -             & CodeWhisperer    \\
    \citet{msai}            & -                 & -             & AI Shell         \\
    \citet{scriptsmith}     & -                 & -             & ScriptSmith      \\
    \citet{bash-assistant}  & -                 & -             & CodeLlama2       \\ \hline
    \textbf{Ours (2025)}    & NL2SH-ALFA        & IC-ALFA       & Llama, Qwen, GPT \\ \hline
  \end{tabular}
  \label{tab:related}
\end{table}

The 2020 NeurIPS NLC2CMD Competition formalized the task of NL2SH translation by providing a human-curated NL2Bash dataset of 9,305 instruction-command pairs and the NL2CMD benchmark for evaluating submitted models \cite{NL2Bash, NL2CMD-Comp}. The competition resulted in numerous NL2SH models and showed that fine-tuning a pre-trained foundation model could outperform dedicated transformer \cite{Magnum}, recurrent neural network \cite{Tellina}, abstract syntax tree \cite{AST}, and sequence to sequence \cite{AInix} based models for the task of NL2SH translation \cite{ShellGPT}.

The NL2CMD benchmarks's FEH parses commands and assigns a similarity score based on the utilities used, order of utilities, and number of utility flags. This heuristic outperforms conventional string comparison techniques, such as edit distance, for measuring the functional equivalence of commands. However, \citet{NL2CMD-Comp} state that the NL2CMD FEH could be improved by executing commands and measuring the similarity of the outputs and side effects. Verification by execution is preferable because Bash is a Turing-complete language, so verifying the equivalence of two commands before execution is undecidable due to side effects \cite{undecided}. Despite this known shortcoming, the NL2CMD benchmark is widely used for model evaluations \cite{NL2CMD}.

\citet{InterCode} address this shortcoming with the InterCode-Bash benchmark. The benchmark uses a subset of 224 instruction-command pairs from the NL2Bash dataset for testing. InterCode's FEH executes the model command and ground truth command in identical Docker containers \cite{docker}. The results of command execution are then compared using three checks. First, the pre and post-execution states of each container are compared using git-diff. Second, the file contents of each container are compared using MD5 hashes. Third, the standard output of both commands are vectorized and compared using the term frequency, inverse document frequency (TFIDF) method \cite{tfidf}. If every check finds the execution results identical, the model and ground truth command are considered functionally equivalent \cite{InterCode}. Although this method is more accurate than previous heuristics, it will fail to identify a valid model command that has syntactically different output from the ground truth command.

\citet{ExeDS} and \citet{vhdl} present similar execution-based frameworks for Jupyter Notebooks and VHDL code, respectively. \citet{exec_based_eval} describe an execution-based framework similar to the InterCode-Bash benchmark using Podman containers. Unfortunately, the code for their FEH and the 50 instruction-command pairs in their test dataset are not public.

\begin{figure*}[t!]
  \centering
  \includegraphics[width=0.99\textwidth]{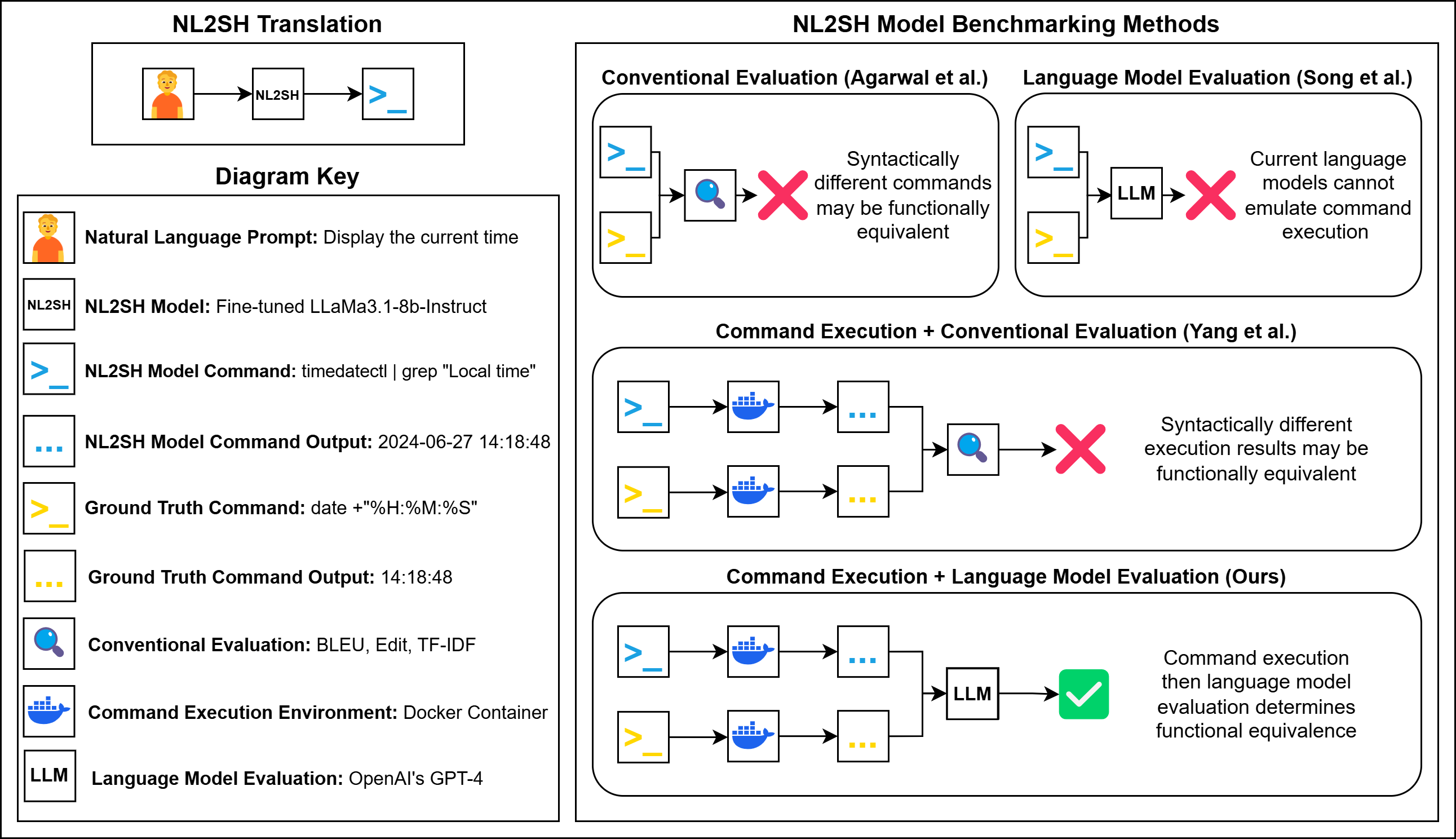}
  \caption{A diagram of NL2SH translation with a comparison of functional equivalence heuristics.}
  \label{fig:icalfa}
\end{figure*}

Focusing on the FEH, \citet{tsed} present a novel benchmark that uses \citeauthor{gpt4}'s GPT-4 model to determine functional equivalence. Their FEH passes the model command and ground truth command to GPT-4 with the prompt, \textit{"Given 2 Bash commands, please generate a similarity score from 0 to 1."} \citet{tsed} find this method fails to determine functional equivalence because current LLMs are unable to emulate command execution. \citet{lackunderstanding} and \citet{limitations} confirm this finding with broader evaluations of LLM's ability to determine semantically equivalent or different pairs of programs. They find LLMs show a lack of depth in understanding code semantics.

\citet{codesift} attempt to advance the work presented by \citet{tsed}, by addressing the scalability constraints of execution-based frameworks. Their FEH, CodeSift, uses an LLM to convert the model command to a natural language description. Then they compare this natural language description with the original natural language task using an LLM. While they find CodeSift to be more effective than conventional FEHs, their work lacks comparison with execution based heuristics. Further, their FEH introduces uncertainty by requiring accurate Bash to natural language translation, which is equally as challenging as the task they aim to measure, natural language to Bash translation. Their follow on work concludes that \textit{"executing [Bash] scripts within a controlled environment would offer more reliable assessments"} \cite{scriptsmith}.

The MultiPL-E benchmark is widely used for evaluating code generation models, containing 540 Bash scripting tasks \cite{multiple}. This benchmark uses unit tests to determine if a generated script produces the expected output for a given input. The use of unit tests is sufficient for this benchmark because its tasks, such as string manipulation and math calculations, are deterministic and result in simple outputs. This method fails to assess file manipulation, system administration and network management tasks because they produce more complex outputs than what can reasonably be assessed with unit tests.

Current SOTA NL2SH models use general purpose LLMs for translation. In practice, users can either accept, reject, or edit the model translation \cite{warp,copilot}. A human-in-the-loop approach is necessary because the models may produce incorrect translations \cite{qwen}. Efforts to improve model performance include fine-tuning and prompt engineering. \citet{finetune} fine-tune the BART, T5, and GPT-2 models on the NL2Bash dataset and find model performance improves as measured by the NL2CMD benchmark. \citet{bash-assistant} conducts a similar study, fine-tuning the CodeLlama2 model on the NL2Bash dataset. Unfortunately, neither of these studies evaluate their models using a reliable execution-based benchmark.

We begin by creating verified and expanded NL2SH datasets starting from multiple datasets presented in previous work. Next, we combine the InterCode execution FEH presented by \citet{InterCode} with the language model evaluation presented by \citet{tsed}. Using our new datasets and FEH, we evaluate methods for improving model performance. Figure \ref{fig:icalfa} summarizes the shortcomings of FEHs in previous work.

\section{Methodology}
\label{sec:methodology}

\subsection{Dataset Creation}
\label{sec:dataset}
Bash is considered a low-resource programming language due to the limited availability of NL2SH data \cite{low-resource}. We aim to augment NL2SH datasets and begin with an evaluation of InterCode. All 224 commands in the InterCode dataset were manually curated from the NL2Bash dataset presented by \citet{NL2Bash}, containing 9,305 instruction-command pairs. The InterCode dataset is significantly smaller than the NL2Bash dataset because a Docker environment is configured for each command, enabling execution. We manually verify all 224 instruction-command pairs and find that over half of the InterCode dataset is erroneous.

Table \ref{tab:errors} shows the number of errors organized by type. We define three types of errors: invalid prompt, invalid command, and invalid environment. An invalid prompt error refers to a natural language instruction that describes an impossible task or does not give enough information to accomplish the task. An invalid command error refers to a Bash command that does not accomplish the task described in the prompt or does not execute. An invalid environment error refers to an incorrect Docker configuration such as a missing file, environment variable, or utility that prevents a valid command from accomplishing the task. Our manual verification reveals 102 instruction-command pairs with one or more errors and 11 duplicate pairs.

\begin{table}[ht!]
  \centering
  \scriptsize
  \caption{InterCode dataset errors.}
  \begin{tabular}{lll}
    \hline
    \textbf{Error Type}         & \textbf{Count} & \textbf{Percentage} \\
    \hline
    Duplicate                   & 11             & 4.9\%               \\
    Invalid Prompt              & 17             & 7.6\%               \\
    Invalid Cmd                 & 24             & 10.7\%              \\
    Invalid Env                 & 18             & 8.0\%               \\
    Invalid Prompt and Cmd      & 29             & 12.9\%              \\
    Invalid Prompt and Env      & 0              & 0.0\%               \\
    Invalid Cmd and Env         & 3              & 1.3\%               \\
    Invalid Prompt, Cmd and Env & 11             & 4.9\%               \\
    \hline
    Invalid Total               & 113            & 50.4\%              \\
    Valid Total                 & 111            & 49.6\%              \\
    \hline
  \end{tabular}
  \label{tab:errors}
\end{table}

We fix 82 of these errors by correcting natural language prompts, Bash commands, and Docker configuration files. We remove 11 duplicate and 20 irreparable pairs from the dataset, resulting in 193 verified pairs. We create 117 more verified pairs by referencing Bash tutorials and books, such as The Linux Command Line by \citet{linux_cli_book} and the Linux Command Line and Shell Scripting Bible by \citet{linux_bible}.

Additionally, for our 300 verified pairs, we create a second Bash command that accomplishes the task described in the prompt. Our final test dataset contains two functionally equivalent, ground truth Bash commands for each natural language instruction, for a total of 600 instruction-command pairs. This is an increase of $441\%$ over the 111 valid commands in the InterCode dataset. Our annotated corrections for the InterCode dataset errors can be found on HuggingFace\footnote{\href{https://huggingface.co/datasets/westenfelder/InterCode-Corrections}{InterCode-Corrections HuggingFace}}.

We collect training data by combining the NL2Bash dataset with three publicly available NL2SH datasets \cite{dataset2, dataset3, dataset4}. Further, we scrape the tldr-pages, a collection of example Bash commands, as a new data source \cite{tldr}. We combine these data sources and deduplicate with exact matching. Then, we use the bashlex parser to remove unparsable commands \cite{bashlex}.

\begin{figure}[ht!]
  \centering
  \includegraphics[width=0.99\columnwidth]{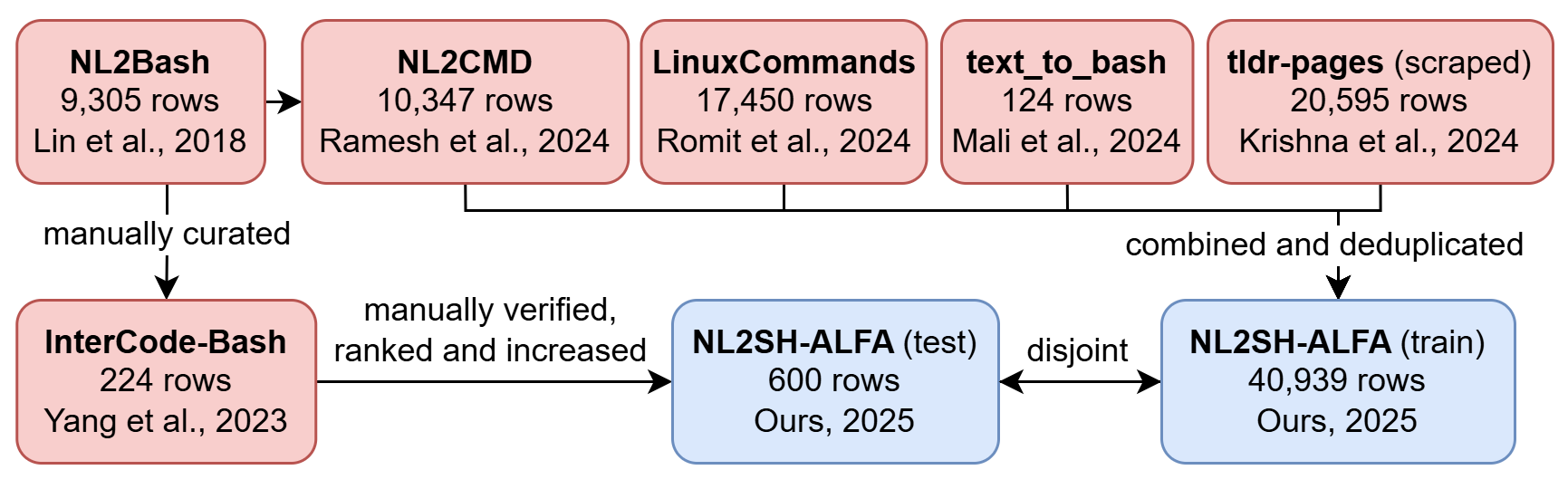}
  \caption{Relationships between NL2SH datasets.}
  \label{fig:datasets}
\end{figure}

We de-conflict our training and test dataset using exact matching and semantic similarity, removing 917 rows from the training data. First, we remove rows from the training data that exactly match instructions or commands in the test data. Next, we remove pairs from the training data with a natural language prompt that is syntactically similar to a prompt in the test data using the mxbai-embed-large-v1 embedding model and a cosine similarity threshold of 0.9 \cite{mxbai}. Our final training dataset contains 40,939 instruction-command pairs, an increase of $135\%$ over the previous largest dataset. Figure \ref{fig:datasets} shows the relationships between data sources used to create our datasets. Our final datasets can be found on HuggingFace\footnote{\href{https://huggingface.co/datasets/westenfelder/NL2SH-ALFA}{NL2SH-ALFA HuggingFace}}.

\subsection{Functional Equivalence Heuristic (FEH)}
Our evaluation of related work in Section \ref{sec:related-work} reveals the InterCode benchmark is more accurate than previous NL2SH benchmarks because its FEH uses execution-based evaluation. However, its TFIDF method for comparing command outputs may fail to determine functional equivalence because syntactically different outputs may convey the same information to the end user.

For example, consider the prompt \textit{"Print the disk usage of the current directory"}, a ground truth command of "\texttt{du -s .}" and a model command of "\texttt{du -d 0 -h}". The first command outputs the number of bytes and the second command outputs the number of bytes in human-readable format. The two commands are functionally equivalent, conditioned on the prompt. However, their outputs contain different characters, resulting in a low similarity score using TfidfVectorizer. Similar issues arise when comparing commands that print hardware information or system time in different formats, display text with line numbers or other delimiters, and use non-deterministic utilities, such as those that interact with the network. The difficulty of determining functional equivalence is exacerbated by ambiguity in natural language prompts, which is an inherent problem with human inputs.

To address this problem, we replace the TfidfVectorizer method for comparing command outputs with an LLM. Our intuition is that an LLM can determine more complex cases of functional equivalence by evaluating the semantics of command outputs with relation to the prompt. Replacing TfidfVectorizer with an LLM increases the computational cost of the FEH. Additionally, since LLMs are stochastic, our FEH has inherent variability. We compare our FEH with previous heuristics in Section \ref{sec:experiments-feh}, finding it achieves superior performance.

We present our FEH and test dataset as a new version of the InterCode benchmark, InterCode-ALFA. Our benchmark and datasets are released under MIT licenses. In addition to the dataset and FEH modifications, we add error handling and update the Docker configuration files to use stable Linux releases. We also identify and fix an error in the benchmark's Docker reset script that causes the filesystem structure of the two execution environments to diverge. We publish the benchmark source code on GitHub\footnote{\href{https://github.com/westenfelder/InterCode-ALFA}{InterCode-ALFA GitHub}} and provide a Python package on PyPI\footnote{\href{https://pypi.org/project/icalfa/}{InterCode-ALFA PyPI}} for ease of use. Our benchmark and dataset can be configured with 10 lines of code, simplifying the process for evaluating new models.

\subsection{Translation Methods}
Using our benchmark, we evaluate the NL2SH performance of the Llama, Qwen and GPT model families \cite{llama3, qwen, gpt4}. We find the models have poor baseline performance and identify three translation failure modes: incorrect output format, incorrect utility and syntactically incorrect Bash command.

Incorrect output format refers to a translation with extraneous information, such as an explanation of the translation, or additional text formatting, such as markdown code blocks. Incorrect utility refers to a translation with a utility that cannot accomplish the task described in the prompt. Syntactically incorrect Bash command refers to a translation that is not valid Bash syntax.

To address these failure modes, we evaluate four methods for improving model performance: markdown parsing \ref{sec:markdown-parser}, constrained decoding (CD) \ref{sec:constrained-decoding}, in-context learning (ICL) \ref{sec:in-context-learning}, and in-weight learning (IWL) \ref{sec:in-weight-learning}. Our results are listed in Section \ref{sec:experiments-methods}.

\subsubsection{Markdown Parser}
\label{sec:markdown-parser}
Despite prompting models with \textit{"You will not output markdown or other formatting"}, translations often include markdown formatting, likely due to instruct tuning. We implement a markdown parser to extract the Bash command from the first code block in model outputs, discarding additional text.

\subsubsection{Constrained Decoding}
\label{sec:constrained-decoding}
We inspect the token probabilities for each ground truth Bash command in our test dataset using the Llama3.1-8b-Instruct model. We find the average relative probability of the first token is four orders of magnitude smaller than the following tokens. In our case, the first token of each command is a Bash utility. This indicates the model is unlikely to select the correct utility as the first token. However, if it does select the correct utility, the following flags and arguments are correct with high probability. We address this by constraining the first tokens of the model output to a list of Bash utilities using grammar-constrained decoding \citet{constrained}.

\subsubsection{In-Context Learning}
\label{sec:in-context-learning}
In-context learning can improve model performance for a variety of tasks \cite{gpt3, icl}. We select 50 representative instruction-command pairs from our training dataset as ICL examples. We create embeddings for the commands using the mxbai-embed-large-v1 model and cluster the embeddings using k-means clustering. The closest instruction-command pair to each centroid is selected as an ICL example. We append these pairs to our translation prompt as show in Figure \ref{fig:prompt-icl}. We evaluate the performance of Llama3.1-8b-Instruct with the number of appended pairs ranging from 1-50 and find the optimal number to be 25, with performance saturating as more pairs are added. We use 25 example instruction-command pairs for all ICL evaluations.

\subsubsection{In-Weight Learning}
\label{sec:in-weight-learning}
We use our training dataset to perform a LoRA fine-tune of the Llama and Qwen models \cite{lora}. We experiment with common hyper-parameters within our hardware constraint of a single Nvidia RTX A6000. We find that training each model for 10 epochs with an adapter rank of 64, adapter alpha of 32, adapter dropout of 0.1, batch size of 32 and learning rate of $1e-5$ results in the best performance. We do not fine-tune the GPT models due to financial constraints and the inability to control training hyper-parameters.

\section{Experiments}
\label{sec:experiments}

\subsection{Evaluation of FEHs}
\label{sec:experiments-feh}
We compare our FEH with the heuristics presented by \citet{tfidf}, \citet{bleu}, \citet{NL2CMD-Comp}, \citet{InterCode} and \citet{tsed} in previous work using our test dataset. A FEH should return true given two functionally equivalent Bash commands, and false given two non-equivalent Bash commands. We record the precision, recall, F1 score and accuracy of each FEH and report our results in Table \ref{tab:feh_results}.

Our test dataset, described in Section \ref{sec:dataset}, is structured \{nl, bash, bash2\}, providing 300 pairs of functionally equivalent commands. To create a set of non-equivalent commands, we arbitrarily rotate the third column of the dataset by ten positions. The result is 600 pairs of Bash commands. Explicitly, each FEH is tested using 300 functionally equivalent pairs $m(t, g(b,e), g(\hat{b},e))=1$ where $g(b, e) \approx g(\hat{b}, e)$ and 300 non-equivalent pairs $m(t, g(b,e), g(\hat{b},e))=0$ where $g(b, e) \neq g(\hat{b}, e)$.

For the bleu and nl2cmd FEHs we use a threshold of 0.75 for functional equivalence. For the tfidf and mxbai-embed FEHs we calculate the cosine similarity of the resulting embeddings and use a threshold of 0.75 for functional equivalence. For the llama-3.1-8b-inst, gpt-3.5-t-0125 and gpt-4-0613 FEHs we pass the tasks and commands to each model using the prompt in Figure \ref{fig:prompt-feh-llm}. For the exec + tfidf and exec + mxbai-embed FEHs, we pass the stdout of command execution to the models, calculate the cosine similarity of the resulting embeddings and use a threshold of 0.75 for functional equivalence. Finally, for the exec + llama-3.1-8b-inst, exec + gpt-3.5-t-0125 and exec + gpt-4-0613 FEHs, we pass the tasks, commands and stdouts of command execution to each model using the prompt in Figure \ref{fig:prompt-feh-exec-llm}. We use a temperature of zero and a static seed value of 123 for all LLMs.

\begin{table}[ht!]
  \centering
  \scriptsize
  \caption{Evaluation of Bash functional equivalence heuristics. Heuristics were tested on a dataset comprising 300 pairs of equivalent commands and 300 pairs of non-equivalent commands. We find execution paired with LLM evaluation significantly increases recall. Bold indicates the highest F1 score and accuracy.}
  \begin{tabular}{llllll}
    \hline
    \textbf{Heuristic}                       & \textbf{Prec.} & \textbf{Rec.} & \textbf{F1}   & \textbf{Acc.} \\
    \hline
    bleu \cite{bleu}                         & 0.99           & 0.39          & 0.56          & 0.69          \\
    nl2cmd \cite{NL2CMD-Comp}                & 0.98           & 0.20          & 0.33          & 0.60          \\ \hline
    tfidf \cite{tfidf}                       & 0.99           & 0.46          & 0.63          & 0.73          \\
    exec + tfidf \cite{InterCode}            & 0.99           & 0.65          & 0.79          & 0.82          \\ \hline
    mxbai-embed   \cite{mxbai}               & 0.84           & 0.88          & 0.86          & 0.85          \\
    exec + mxbai-embed \textbf{(Ours)}       & 0.97           & 0.83          & 0.90          & 0.90          \\ \hline
    llama-3.1-8b-inst \cite{llama3}          & 1.00           & 0.05          & 0.10          & 0.53          \\
    exec + llama-3.1-8b-inst \textbf{(Ours)} & 0.88           & 0.74          & 0.80          & 0.82          \\ \hline
    gpt-3.5-t-0125  \cite{gpt3}              & 1.00           & 0.37          & 0.54          & 0.69          \\
    exec + gpt-3.5-t-0125 \textbf{(Ours)}    & 0.98           & 0.60          & 0.75          & 0.80          \\ \hline
    gpt-4-0613 \cite{tsed}                   & 1.00           & 0.51          & 0.68          & 0.76          \\
    exec + gpt-4-0613 \textbf{(Ours)}        & 0.99           & 0.91          & \textbf{0.95} & \textbf{0.95} \\
    \hline
  \end{tabular}
  \label{tab:feh_results}
\end{table}

\subsection{Evaluation of Translation Methods}
\label{sec:experiments-methods}
We evaluate the impact of parsing, constrained decoding, in-context learning and in-weight learning on the NL2SH performance of the Llama, Qwen and GPT model families. All models are evaluated using version 0.3.6 of the InterCode-ALFA benchmark with the execution + mxbai-embed FEH. Our results are summarized in Table \ref{tab:model_results}.

We use the prompt in Figure \ref{fig:prompt-baseline} for the baseline evaluation. For the constrained decoding evaluation, we use the prompt in Figure \ref{fig:prompt-other} and constrain the first tokens of the model output to a list of Bash utilities, as described in Section \ref{sec:constrained-decoding}. For the parser evaluation, we use the prompt in Figure \ref{fig:prompt-other} and pass model outputs to a markdown parser, as described in Section \ref{sec:markdown-parser}. For the in-context learning evaluation, we use the prompt in Figure \ref{fig:prompt-icl}. Finally, for the in-weight learning evaluation, we use the prompt in Figure \ref{fig:prompt-other} and the fine-tuned models described in Section \ref{sec:in-weight-learning}. We use a temperature of zero and a static seed value of 123 for all model evaluations.

\begin{table}[ht!]
  \centering
  \scriptsize
  \caption{Impact of constrained decoding, parsing, in-context learning and in-weight learning on the NL2SH performance of Llama, Qwen and GPT models. Accuracy is measured using the exec + mxbai-embed FEH. Translation method can improve performance up to 32\% over the baseline, but model size remains the dominant factor. The highest accuracy in each row and column is indicated with bold and an underline, respectively.}
  \begin{tabular}{llllll}
    \hline
    \textbf{Model}              & \textbf{Base}             & \textbf{CD}      & \textbf{Parse}   & \textbf{ICL}              & \textbf{IWL}     \\
    \hline
    llama-3.2-1b-instruct       & 0.12                      & 0.19             & 0.32             & 0.34                      & \textbf{0.37}    \\
    llama-3.2-3b-instruct       & 0.17                      & 0.39             & \textbf{0.49}    & 0.47                      & 0.47             \\
    llama-3.1-8b-instruct       & 0.46                      & \underline{0.51} & 0.53             & \textbf{0.56}             & 0.40             \\ \hline
    qwen2.5-coder-0.5b-instruct & 0.10                      & 0.05             & 0.35             & \textbf{0.36}             & 0.27             \\
    qwen2.5-coder-1.5b-instruct & 0.21                      & 0.06             & \textbf{0.50}    & 0.44                      & 0.19             \\
    qwen2.5-coder-3b-instruct   & 0.26                      & 0.06             & \textbf{0.58}    & 0.50                      & \underline{0.51} \\
    qwen2.5-coder-7b-instruct   & 0.61                      & 0.08             & \textbf{0.62}    & 0.62                      & \underline{0.51} \\ \hline
    gpt-3.5-turbo-0125          & 0.58                      & -                & 0.67             & \textbf{0.69}             & -                \\
    gpt-4o-mini-2024-07-18      & 0.71                      & -                & \textbf{0.72}    & 0.71                      & -                \\
    gpt-4o-2024-08-06           & \textbf{\underline{0.74}} & -                & \underline{0.73} & \underline{0.73}          & -                \\
    gpt-4-0613                  & 0.68                      & -                & 0.68             & \textbf{\underline{0.73}} & -                \\
    \hline
  \end{tabular}
  \label{tab:model_results}
\end{table}

\section{Discussion}
\label{sec:discussion}
\subsection{Dataset}
Our first research question aims to validate NL2SH datasets to ensure models are evaluated using valid translations. Manual verification of the InterCode dataset identified over half of the instruction-command pairs as erroneous. We find human verification of data is important for reliable evaluations.

Manual creation and verification of our test dataset took over 100 hours, highlighting the need for more efficient means to verify larger datasets. Further, since the InterCode dataset is sampled from the NL2Bash dataset, there is a risk the NL2Bash dataset contains a significant percentage of erroneous data. This is concerning because the NL2Bash dataset is commonly used to train NL2SH models \cite{Magnum,Tellina,ShellGPT,AST,bash-assistant}.

We are confident our training dataset contains valid data due to our filtering process to remove invalid Bash commands. Moreover, fine-tuning the Llama and Qwen models using our dataset results in an average performance increase of $11\%$.

\subsection{Functional Equivalence Heuristic}
Our second research question aims to design an FEH that accurately measures the quality of model translations. We find that command execution paired with LLM evaluation of command outputs can determine the functional equivalence of Bash commands with $95\%$ accuracy. The ability of LLMs to condition command outputs on natural language inputs is an advancement that was not possible with previous heuristics. Further, command execution improves performance across methods and the use of an LLM significantly increases recall. Broadly, LLM evaluation of execution outputs is a promising advancement for measuring the functional correctness of generated code and more investigation is warranted.

In accordance with \citet{lackunderstanding}, we find that without execution, current LLMs are poor arbiters of functional equivalence, achieving similar performance when compared to conventional evaluation methods. Non-execution methods likely fail because two commands can be syntactically similar and yield different results when executed. For example, changing a single flag can result in vastly different command outputs. Further, two commands with no syntactic similarity can yield identical results when executed. For example, the \texttt{awk} and \texttt{sed} utilities can accomplish identical text processing tasks but use different domain-specific languages, requiring different syntax. Notably, the low recall of bleu and nl2cmd FEHs indicates these methods cannot identify cases where syntactically different commands are functionally equivalent.

\subsection{Translation Methods}
Our third research question aims to improve the accuracy of NL2SH models as measured by a reliable benchmark. We find that constrained decoding, parsing, in-context learning and in-weight learning can improve model performance by up to $32\%$. Our baseline evaluation shows that model performance is correlated with number of parameters.

We find that constrained decoding is model dependent, with performance increases for Llama models and significant performance decreases for Qwen models. Parsing and ICL provide performance increases across Llama and Qwen models, with average increases of $21\%$ and $19\%$, respectively. However, these methods have a decreasing impact as model size increases. This is evidence that incorrect output format is the dominant failure mode for small (less than 7b parameter) models.

With IWL, llama-3.2-3b-instruct and qwen2.5-coder-0.5b-instruct achieve the baseline performance of llama-3.1-8b-instruct and qwen2.5-coder-3b-instruct, respectively. Despite performance gains for small models, fine-tuning decreases the performance of llama-3.1-8b-instruct and qwen2.5-coder-7b-instruct. This is likely due to computational constraints on the size of our LoRA adapters, which we are unable to scale with model size.

We find that gpt-4o-2024-08-06 achieves SOTA performance on our benchmark, correctly translating $74\%$ of test cases. From the total of our experiences, we find that NL2SH translation is a difficult task for current models, necessitating improvements before models can be used in practice.

\section{Conclusion}
\label{sec:conclusion}
In this paper, we explore applications for LLMs in NL2SH translation and benchmarking. We identify issues with current benchmarks, including inaccurate datasets and unreliable functional equivalence heuristics. To address these problems, we correct and expand NL2SH datasets and create a new heuristic to determine the functional equivalence of Bash commands. We assess our heuristic and find that Bash command execution paired with language model evaluation of command outputs can determine the functional equivalence of commands more accurately than previous heuristics. Using our dataset and heuristic, we evaluate how constrained decoding, parsing, in-context learning and in-weight learning impact the performance of Llama, Qwen and GPT models. We find that parsing and in-context learning reliably improve the performance of open and closed-source LLMs for the task of NL2SH translation. Ultimately, we find that NL2SH translation is a difficult task for LLMs, necessitating further research in this field. In future work, we plan to investigate efficient means to verify our training dataset and conduct more fine-tuning experiments.

\section{Limitations}
\label{sec:limitations}
This work presents a verified and expanded NL2SH test dataset. However, due to the time and effort required to manually configure an execution environment for each command, the dataset remains small, with only 600 test cases. In contrast, our training dataset is too large for manual verification, and we are unable to guarantee its correctness. Our datasets are limited to English prompts and one-line Bash commands. We do not consider other natural languages or scripting languages.

Although improved over previous methods, our functional equivalence heuristic has inherent variability due to the use of an LLM, requiring multiple runs to assess model performance. The use of an LLM also increases the computational cost of running our heuristic compared to conventional methods. Finally, despite improved model performance with constrained decoding, parsing, ICL and IWL, the accuracy of SOTA LLMs for NL2SH translation remains low, motivating further research.

\section{Ethical Considerations}
\label{sec:ethics}
Due to the low performance of current NL2SH models, using these models in practice could result in invalid commands that have unintended effects on a system. We recommend that model-generated commands are never used without human verification. Further, we recommend that users test commands using a sand-boxed environment, such as try \cite{try}, before running them on personal systems. Figure \ref{fig:bricked} shows an example of a dangerous command observed during model testing. The hallucinated \texttt{rm -f /dev/null} command corrupted our benchmark's Docker container. While our benchmark automatically creates a new Docker container to handle this type of error, the command could have corrupted a user's system.

\begin{figure}[ht!]
  \centering
  \small
  \begin{tabular}{l}
    \hline
    \textbf{Natural Language Prompt:}                      \\
    \textit{Recursively remove all empty folders from the} \\
    \textit{/system/temp folder.}                          \\
    \hline
    \textbf{Ground Truth Command:}                         \\
    \texttt{find /system/temp -type d -empty -delete}      \\
    \hline
    \textbf{Llama3.1-8b-Instruct Command:}                 \\
    \texttt{find /system/temp -type d -empty -delete}      \\
    \texttt{-print; rm -f /dev/null 2>\&1}                 \\
    \hline
  \end{tabular}
  \caption{Dangerous translation observed in testing.}
  \label{fig:bricked}
\end{figure}

Natural language to Bash translation aims to increase computer accessability by simplifying interactions with the command-line interface. Unfortunately, good and bad actors can benefit from increased accessability. Models could be used to generate malicious Bash commands. This risk is difficult to mitigate because malicious use depends on the intent of the user. For example, a command to delete files could be used for a legitimate or harmful purpose. We do not believe current NL2SH models pose any risks beyond those of other readily available Bash resources.

\bibliography{paper}

\onecolumn
\appendix{}
\section*{Appendix A. Prompts}
\label{sec:appendix}

\begin{figure*}[ht!]
  \centering
  \small
  \begin{tabular}{l}
    \hline
    \textbf{Functional Equivalence Heuristic Prompt: LLM} \\
    \hline
    \\
    \begin{minipage}{0.97\textwidth}
      \textit{You will be given a task and two Bash commands. The first command is the ground truth. If the second command accomplishes the task, return true. Otherwise, return false. Only output 'true' or 'false'. Task: } \texttt{natural\allowbreak\_language\allowbreak\_prompt}\textit{, Ground Truth Command: }\texttt{ground\allowbreak\_truth\allowbreak\_command}\textit{, Model Command: }\texttt{model\allowbreak\_command}\textit{.}
    \end{minipage}
    \\\\
    \hline
  \end{tabular}
  \caption{Prompt for evaluating the functional equivalence of Bash commands.}
  \label{fig:prompt-feh-llm}
\end{figure*}

\begin{figure*}[ht!]
  \centering
  \small
  \begin{tabular}{l}
    \hline
    \textbf{Functional Equivalence Heuristic Prompt: Execution + LLM} \\
    \hline
    \\
    \begin{minipage}{0.97\textwidth}
      \textit{You will be given a task, two Bash commands, and the output of the two Bash commands. The first command is the ground truth. If the second command accomplishes the task, return true. Otherwise, return false. Only output 'true' or 'false'. Task: } \texttt{natural\allowbreak\_language\allowbreak\_prompt}\textit{, Ground Truth Command: }\texttt{ground\allowbreak\_truth\allowbreak\_command}\textit{, Model Command: }\texttt{model\allowbreak\_command}\textit{, Ground Truth Command Output: }\texttt{ground\allowbreak\_truth\allowbreak\_command\allowbreak\_output}\textit{, Model Command Output: }\texttt{model\allowbreak\_command\allowbreak\_output}\textit{.}
    \end{minipage}
    \\\\
    \hline
  \end{tabular}
  \caption{Prompt for evaluating the functional equivalence of Bash commands after execution. Note the addition of command outputs compared to the prompt in Figure \ref{fig:prompt-feh-llm}.}
  \label{fig:prompt-feh-exec-llm}
\end{figure*}

\begin{figure*}[ht!]
  \centering
  \small
  \begin{tabular}{l}
    \hline
    \textbf{Translation Prompt: Baseline} \\
    \hline
    \\
    \begin{minipage}{0.97\textwidth}
      \textit{Your task is to translate a natural language instruction to a Bash command. You will receive an instruction in English and output a Bash command that can be run in a Linux terminal. You will not output markdown or other formatting. You will not include additional information. }\texttt{natural\_language\_prompt}
    \end{minipage}
    \\\\
    \hline
  \end{tabular}
  \caption{NL2SH translation prompt used in the baseline evaluation.}
  \label{fig:prompt-baseline}
\end{figure*}

\begin{figure*}[ht!]
  \centering
  \small
  \begin{tabular}{l}
    \hline
    \textbf{Translation Prompt: Parser, Constrained Decoding and In-Weight Learning} \\
    \hline
    \\
    \begin{minipage}{0.97\textwidth}
      \textit{Your task is to translate a natural language instruction to a Bash command. You will receive an instruction in English and output a Bash command that can be run in a Linux terminal. }\texttt{natural\_language\_prompt}
    \end{minipage}
    \\\\
    \hline
  \end{tabular}
  \caption{NL2SH translation prompt used in the parsing, constrained decoding and in-weight learning evaluations.}
  \label{fig:prompt-other}
\end{figure*}

\begin{figure*}[ht!]
  \centering
  \scriptsize
  \begin{tabular}{l}
    \hline
    \textbf{Translation Prompt: In-Context Learning} \\
    \hline
    \\
    \begin{minipage}{0.97\textwidth}
      \textit{Your task is to translate a natural language instruction to a Bash command. You will receive an instruction in English and output a Bash command that can be run in a Linux terminal.\\\\
        Show logged-in users info\\
        w\\\\
        Print the contents of "xx.sh"\\
        cat xx.sh\\\\
        Change owner to "root" and group to "www-data" of "/foobar/test\_file"\\
        chown root:www-data /foobar/test\_file\\\\
        delete all the text files in the current folder\\
        find . -type f -name "*.txt" -delete\\\\
        find all the files in the /path folder and delete them\\
        find /path -type f -delete\\\\
        Print the exit status of the last executed command\\
        echo \$?\\\\
        Display a tree of processes\\
        pstree\\\\
        Display information about all CPUs\\
        lscpu\\\\
        Make an HTTPS GET request to example.com and dump the contents in `stdout`\\
        curl https://example.com\\\\
        Display system memory\\
        free\\\\
        List all files, including hidden files\\
        ls -a\\\\
        Print a sequence from 1 to 10\\
        seq 10\\\\
        Get the properties of all the user limits\\
        ulimit -a\\\\
        List the name and status of all services\\
        service --status-all\\\\
        Display a calendar for the current month\\
        cal\\\\
        Show the environment\\
        env\\\\
        create directory TestProject\\
        mkdir TestProject\\\\
        Query the default name server for the IP address of example.com\\
        nslookup example.com\\\\
        Print Hello World\\
        echo "Hello World"\\\\
        List all bound commands and their hotkeys\\
        bind -p\\\\
        Display the openssl version\\
        openssl version\\\\
        Print current time, uptime, number of logged-in users\\
        uptime\\\\
        Print file system disk space usage\\
        df\\\\
        List all configuration values available\\
        getconf -a\\\\
        Delete empty folder 'nonsense\_dir'.\\
        rmdir nonsense\_dir\\\\
      }
      \texttt{natural\_language\_prompt}
    \end{minipage}
    \\\\
    \hline
  \end{tabular}
  \caption{NL2SH translation prompt used in the in-context learning evaluation.}
  \label{fig:prompt-icl}
\end{figure*}

\end{document}